\documentclass[tikz]{article}

\usepackage{PRIMEarxiv}

\usepackage[utf8]{inputenc} 
\usepackage[T1]{fontenc}    
\usepackage{hyperref}       
\usepackage{url}            
\usepackage{booktabs}       
\usepackage{amsfonts}       
\usepackage{nicefrac}       
\usepackage{microtype}      
\usepackage{lipsum}
\usepackage{fancyhdr}       
\usepackage{graphicx}       
\usepackage{csquotes}
\usepackage{mathtools} 
\usepackage{booktabs} 
\usepackage{algorithm}
\usepackage[noend]{algorithmic}
\usepackage{multirow}
\usepackage{caption}
\usepackage{subcaption}
\usepackage{amssymb}
\usepackage{graphicx}
\usepackage{xfrac}
\usepackage{xcolor}
\usepackage{ulem}
\usepackage{microtype}
\usepackage{hyperref}
\usepackage[acronym]{glossaries}

\newacronym{ml}{ML}{Machine Learning}
\newacronym{dl}{DL}{Deep Learning}
\newacronym{gnn}{GNN}{Graph Neural Network}
\newacronym{gcn}{GCN}{Graph Convolutional Network}
\newacronym{kmean}{K-Mean}{K-Means Clustering}
\newacronym{pca}{PCA}{Principal Component Analysis}
\newacronym{umap}{UMAP}{Uniform Manifold Approximation and Projection}
\newacronym{tsne}{t-SNE}{t-distributed Stochastic Neighbor Embedding}
\newacronym{rf}{RF}{Random Forest}
\newacronym{auc-roc}{AUC-ROC}{Area Under the Receiver Operating Characteristic Curve}
\newacronym{api}{API}{Application Programming Interface} 
\newacronym{hdb}{HDBSCAN}{Hierarchical Density-Based Spatial Clustering of Applications with Noise}

\usepackage{cleveref}

\title{
    Deep Learning for School Dropout Detection: A Comparison of Tabular and Graph-Based Models for Predicting At-Risk Students
}

\author{
Pablo G. Almeida$^1$, Guilherme A. L. Silva$^1$, Valéria Santos$^1$, Gladston Moreira$^1$\\ 
\textbf{Pedro Silva$^1$ and Eduardo Luz}$^1$ \\
 $^1$Computing Department, Universidade Federal de Ouro Preto, Ouro
Preto, 35402-136, Minas Gerais, Brazil.\\
    \texttt{eduluz@ufop.edu.br}
}

\begin{document}
\maketitle

\begin{abstract}
Student dropout is a significant challenge in educational systems worldwide, leading to substantial social and economic costs. Predicting students at risk of dropout allows for timely interventions. While traditional \gls{ml} models operating on tabular data have shown promise, \glspl{gnn} offer a potential advantage by capturing complex relationships inherent in student data if structured as graphs. This paper investigates whether transforming tabular student data into graph structures, primarily using clustering techniques, enhances dropout prediction accuracy. We compare the performance of \glspl{gnn} (a custom \gls{gcn} and GraphSAGE) on these generated graphs against established tabular models (\gls{rf}, XGBoost, and TabNet) using a real-world student dataset. Our experiments explore various graph construction strategies based on different clustering algorithms (K-Means, HDBSCAN) and dimensionality reduction techniques (\gls{pca}, \gls{umap}). Our findings demonstrate that a specific \gls{gnn} configuration, GraphSAGE on a graph derived from PCA-KMeans clustering, achieved superior performance, notably improving the macro F1-score by approximately 7 percentage points and accuracy by nearly 2 percentage points over the strongest tabular baseline (XGBoost). However, other \gls{gnn} configurations and graph construction methods did not consistently surpass tabular models, emphasizing the critical role of the graph generation strategy and \gls{gnn} architecture selection. This highlights both the potential of \glspl{gnn} and the challenges in optimally transforming tabular data for graph-based learning in this domain.
\end{abstract}

\keywords{Student Dropout Prediction \and Graph Neural Networks \and Tabular Data \and Machine Learning \and Educational Data Mining \and Clustering.}

\section{Introduction}
\label{sec:introduction}


Student dropout represents a critical issue across all educational levels, with particularly alarming rates in higher education in countries like Brazil, where it can reach 57\% in public and private institutions \footnote{\url{https://www.correiobraziliense.com.br/euestudante/ensino-superior/2024/05/6852929-ensino-superior-no-brasil-tem-57-de-evasao-na-rede-publica-e-privada.html}}. This phenomenon not only curtails individual student potential but also incurs substantial financial losses for educational institutions \cite{silva2022evasao} and wider socio-economic consequences. Early identification of students at risk of dropout is paramount, as it enables institutions to implement targeted interventions and support mechanisms.

Predicting student dropout typically involves analyzing diverse data, often available in tabular format, encompassing demographic information, academic performance, and engagement metrics. While traditional \gls{ml} models like \gls{rf} \cite{breiman2001random} and XGBoost \cite{friedman2000additive}, as well as \gls{dl} models for tabular data like TabNet \cite{arik2020tabnet}, have achieved considerable success \cite{cambruzzi2015dropout}, they may not fully capture latent, complex interdependencies between students or student attributes. 


\Glspl{gnn} \cite{li2024graph} have emerged as powerful tools for learning from graph-structured data, offering the potential to model such relationships more effectively. However, educational data is not inherently graph-structured. Converting tabular data into meaningful graph representations that \glspl{gnn} can leverage poses a significant challenge \cite{hofer2020graph, somepalli2021saint}. The central research question (RQ1) we address is: \textit{Can transforming tabular student data into graph structures via clustering techniques improve the accuracy of dropout prediction models compared to state-of-the-art tabular methods?} Furthermore (RQ2): \textit{How do different graph construction strategies, based on various clustering and dimensionality reduction combinations, affect \gls{gnn} performance in this task?}


This paper presents an empirical study to address these questions by proposing and evaluating a comprehensive pipeline. Initially, a real-world student dataset undergoes thorough preprocessing. Subsequently, we employ clustering algorithms, such as K-Means and HDBSCAN, often in conjunction with dimensionality reduction techniques including \gls{pca}, \gls{umap}, and \gls{tsne}, to group students based on their feature similarity; the quality of these groupings is assessed using the Silhouette Score. Following this, graph representations are constructed from the identified clusters, where nodes correspond to student instances and edges signify co-membership within clusters or proximity in the reduced feature space. These generated graphs then serve as input for training and evaluating \gls{gnn} models, specifically a custom \gls{gcn} and GraphSAGE, for the task of predicting student dropout status (Enrolled, Dropout, or Graduate). Finally, the performance of these \gls{gnn} models is systematically compared against strong baselines that operate directly on the original tabular data, namely \gls{rf}, XGBoost, and TabNet.


Our experiments, conducted using a publicly available dataset \cite{predict_students'_dropout_and_academic_success_697}, demonstrate that certain configurations of graph neural networks (GNNs) applied to cluster-derived graphs can attain performance levels comparable to, and in one case marginally surpassing, those of traditional tabular models. However, these performance gains are not consistently significant across all employed graph construction techniques.
This suggests that simple clustering-based approaches may not always unlock the full potential of \glspl{gnn} for tabular educational data, and the heterogeneity of such data presents ongoing challenges \cite{somepalli2021saint}.

The main contributions of this work are:
\begin{itemize}
    \item A systematic investigation into the efficacy of converting tabular educational data to graph structures using clustering for student dropout prediction.
    \item An empirical comparison of \gls{gnn} models against established tabular \gls{ml} and \gls{dl} methods on this specific task and dataset.
    \item An analysis of how different clustering (K-Means, HDBSCAN) and dimensionality reduction (\gls{pca}, \gls{umap}) combinations for graph construction impact predictive performance.
    \item Insights into the challenges and potential limitations of current graph representation learning techniques for heterogeneous tabular educational data.
\end{itemize}


\section{Related Work}
\label{sec:related_work}


The prediction of student dropout has been extensively studied using traditional \gls{ml} techniques on tabular data. Cambruzzi et al. \cite{cambruzzi2015dropout} utilized learning analytics trails, while many other studies have employed models like decision trees, support vector machines, \gls{rf} \cite{breiman2001random}, and XGBoost \cite{friedman2000additive} with varying degrees of success, often highlighting the importance of feature engineering and handling imbalanced data. More recently, \gls{dl} architectures for tabular data, such as TabNet \cite{arik2020tabnet} and TabTransformer \cite{huang2020tabtransformertabulardatamodeling}, have shown competitive performance by learning feature representations automatically. Our work contrasts by exploring the transformation of tabular data into graphs.

The application of \glspl{gnn} to tabular data is an emerging research area. Li et al. \cite{li2024graph} provides a comprehensive survey, categorizing methods for constructing graphs from tabular data. Common approaches include k-Nearest Neighbors (k-NN) graphs, fully connected graphs with learned edge weights, or graphs derived from domain knowledge. Our focus on explicit clustering as a primary mechanism for graph construction differs from many existing approaches that rely on direct similarity metrics or assume latent graph structures.

Several studies have applied \glspl{gnn} in educational contexts. Xiaochen et al. \cite{pmlr-v245-xiaochen24a} introduced a \gls{gnn} model using structural density-based sampling, leveraging Jaccard coefficient and Pearson correlation to capture similarity between students for performance prediction. While related, their graph construction does not explicitly employ hierarchical clustering. Li et al. \cite{Li2022Study} proposed a pipeline building multiple graphs with distinct topologies based on similarity metrics and integrating information via attention for student performance prediction. Our work, in contrast, focuses on a direct comparison resulting from cluster-based graph conversion against purely tabular methods.

In \cite{hofer2020graph}, authors explore Graph Filtration Learning to create graphs and then filtering them, which is related to the idea of constructing meaningful graph structures. Musa et al. \cite{MUSA2025103092} use graph-based approaches for integrating heterogeneous learning analytics data via ontologies. The work proposed in \cite{Ifenthaler_Gibson_Dobozy_2018} applies network graph analysis to inform learning design, but not for direct dropout prediction.
Our contribution lies in the specific investigation of clustering-based conversion for dropout prediction and the direct, comparative evaluation against state-of-the-art tabular models, which is not extensively covered in existing literature for this particular problem formulation.

\section{Background}
\label{sec:background}


\subsection{Student Dropout Prediction}
Student dropout is the premature cessation of studies before course completion. Formally, given a set of students \(S = \{s_1, s_2, \dots, s_N\}\), each student \(s_i\) is represented by a feature vector \( \mathbf{x}_i \in \mathbb{R}^D \), where \(D\) is the number of features (e.g., demographics, academic history, engagement metrics). Each student also has a label \(y_i \in \mathcal{Y}\) indicating their status, where \(\mathcal{Y} = \{\text{Enrolled, Dropout, Graduate}\}\) in our case. The objective is to learn a predictive function \(f: \mathbb{R}^D \rightarrow \mathcal{Y}\) that accurately maps student features to their dropout status.

\subsection{Tabular Learning Models}
Traditional machine learning models operate directly on these tabular feature vectors \( \mathbf{x}_i \). In this work, we will evaluate two traditional machine learning algorithms:
\begin{itemize}
    \item \textbf{\Gls{rf}} \cite{breiman2001random}: An ensemble learning method that constructs multiple decision trees during training and outputs the mode of the classes (classification) or mean prediction (regression) of the individual trees.
    \item \textbf{XGBoost} \cite{chen2016xgboost}: A gradient boosting framework that builds trees sequentially, where each new tree corrects errors made by previously trained trees. It is known for its high performance and regularization capabilities.
\end{itemize}

We also evaluate a model that is based on Deep Learning. The evaluated architecture is:
\begin{itemize}
    \item \textbf{TabNet} \cite{arik2020tabnet}: Architecture designed for tabular data. It uses sequential attention to choose which features to reason from at each decision step, enabling interpretability and efficient learning.
\end{itemize}

\subsection{Graph Neural Networks (GNNs)}
\Glspl{gnn} are a class of neural networks designed to operate directly on graph-structured data \cite{wu2020comprehensive}. A graph is denoted as \(\mathcal{G} = (\mathcal{V}, \mathcal{E})\), where \(\mathcal{V}\) is the set of nodes (or vertices) and \(\mathcal{E}\) is the set of edges. Each node \(v \in \mathcal{V}\) can have features \(\mathbf{h}_v \in \mathbb{R}^{F_v}\). \Glspl{gnn} typically learns node representations by aggregating information from their neighbors. In this work, we evaluate two main architectures based on graphs:
\begin{itemize}
    \item \textbf{\Gls{gcn}} \cite{kipf2017semi}: A type of \gls{gnn} that learns node representations by aggregating feature information from their local graph neighborhoods using a spectral approach, simplified to a spatial message-passing scheme. For a layer \(l\), the representation \(\mathbf{h}_v^{(l+1)}\) for node \(v\) is:
    \begin{equation}
        \mathbf{h}_v^{(l+1)} = \phi\left( \sum_{u \in \mathcal{N}(v) \cup \{v\}} \frac{1}{\sqrt{\deg(v)\deg(u)}} \mathbf{W}^{(l)} \mathbf{h}_u^{(l)} \right)
        \label{eq:gcn}
    \end{equation}
    where \(\mathcal{N}(v)\) is the set of neighbors of node \(v\), \(\deg(v)\) is the degree of node \(v\), \(\mathbf{W}^{(l)}\) is a trainable weight matrix for layer \(l\), and \(\phi\) is an activation function. 
    \item \textbf{GraphSAGE} \cite{hamilton2017inductive}: An inductive \gls{gnn} framework that learns functions to generate node embeddings by sampling and aggregating features from a node's local neighborhood. For a node \(v\) and layer \(l\):
    \begin{equation}
        \mathbf{h}_{\mathcal{N}(v)}^{(l+1)} = \text{AGGREGATE}^{(l+1)} \left( \{ \mathbf{h}_u^{(l)}, \forall u \in \mathcal{N}(v) \} \right)
        \label{eq:graphsage_agg}
    \end{equation}
    \begin{equation}
        \mathbf{h}_v^{(l+1)} = \phi \left( \mathbf{W}^{(l+1)} \cdot \text{CONCAT}(\mathbf{h}_v^{(l)}, \mathbf{h}_{\mathcal{N}(v)}^{(l+1)}) \right)
        \label{eq:graphsage_update}
    \end{equation}
    where AGGREGATE can be a mean, pooling, or LSTM aggregator.
\end{itemize}

\subsection{Clustering and Dimensionality Reduction}
To convert tabular student data into a graph-based representation—with nodes and connections—we first apply dimensionality reduction to project each student instance into a lower-dimensional space. Then, we group similar instances using clustering techniques, enabling the identification of meaningful structures that can later be represented as graph edges. 
\begin{itemize}
    \item \textbf{\Gls{kmean}}: An iterative algorithm that partitions \(N\) observations into \(K\) clusters by minimizing the within-cluster sum of squares.
    \cite{macqueen1967some}
    \item \textbf{\Gls{hdb}}: A hierarchical, density-based clustering algorithm that extends DBSCAN by converting it into a hierarchical clustering algorithm, then extracting flat clusters based on stability.
    \cite{campello2013density}
    \item \textbf{\Gls{pca}}: A linear dimensionality reduction technique that transforms data into a new coordinate system such that the greatest variance by some scalar projection of the data lies on the first coordinate (the first principal component), the second greatest variance on the second coordinate, and so on.
    \cite{hotelling1933analysis}
    \item \textbf{\Gls{umap}}: A non-linear dimensionality reduction technique that constructs a high-dimensional graph representation of the data, then optimizes a low-dimensional graph to be as structurally similar as possible.\cite{mcinnes2020umapuniformmanifoldapproximation}
    \item
    \textbf{\Gls{tsne}}:A non-linear dimensionality reduction technique that models pairwise similarities between points in high dimensions and maps them to a low-dimensional space, preserving local structure for visualization.
    \cite{JMLR:v9:vandermaaten08a}
    
\end{itemize}

\subsection{Problem Setting and Notation}
\label{subsec:problem_setting}
Let \(\mathcal{D}_{tab} = \{(\mathbf{x}_i, y_i)\}_{i=1}^N\) be the initial tabular dataset, where \(\mathbf{x}_i \in \mathbb{R}^D\) is the feature vector for student \(i\) and \(y_i \in \mathcal{Y}\) is their corresponding dropout status.
Our primary hypothesis (H1) is that converting \(\mathcal{D}_{tab}\) into a graph \(\mathcal{G} = (\mathcal{V}, \mathcal{E})\), where \(\mathcal{V} = \{v_1, \dots, v_N\}\) represents students and edges \(\mathcal{E}\) encode relationships derived from clustering, allows \glspl{gnn} to achieve superior predictive performance compared to models operating on \(\mathcal{D}_{tab}\).

The graph construction process involves:
1.  Optional dimensionality reduction: \( \mathbf{x}_i \rightarrow \mathbf{x}'_i \in \mathbb{R}^{D'} \), where \(D' < D\).
2.  Clustering: Assign each student \(\mathbf{x}'_i\) (or \(\mathbf{x}_i\)) to a cluster \(c_k \in \{C_1, \dots, C_K\}\).
3.  Edge creation: An edge \((v_i, v_j) \in \mathcal{E}\) exists if students \(s_i\) and \(s_j\) belong to the same cluster \(c_k\). Variations can include connecting students if they are close in the reduced latent space (e.g., using KD-Trees). Node features for \(v_i\) in \(\mathcal{G}\) are the original \(\mathbf{x}_i\).

The task is then to train a \gls{gnn} model \(f_{GNN}(\mathcal{G}, \mathbf{X})\) to predict \(y_i\) for nodes \(v_i\), where \(\mathbf{X}\) is the matrix of node features. We compare \(f_{GNN}(\mathbf{X})\) against tabular models \(f_{tab}(\mathbf{X})\). This directly addresses RQ1 and RQ2.

\section{Experimental Methodology}
\label{sec:methodology}


This section details our approach for converting tabular student data into graph structures and applying \glspl{gnn} for dropout prediction, along with the setup for baseline tabular models.

\subsection{Data Acquisition and Preprocessing}
The primary dataset used in this study is from the  ``Predict Students' Dropout and Academic Success'' dataset~\cite{martins2021early} \footnote{\url{https://archive.ics.uci.edu/dataset/697/predict+students+dropout+and+academic+success}}. It contains anonymized features related to student demographics, socio-economic background, and academic performance for a higher education institution, with the target variable indicating whether a student is "Enrolled", "Dropout", or "Graduate".

The preprocessing steps starts with handling missing data by checking for null values using \texttt{data\_df.isnull().sum()}; columns without missing data were retained, though imputation (e.g., by mean, median, or mode) would have been considered if necessary. Subsequently, header inconsistencies were corrected using \texttt{data\_df.rename(columns=\{\dots\}, inplace=True)} to standardize names like “Nacionality” to “Nationality” and “Daytime/evening attendance\\t” to “Daytime attendance”. Categorical feature encoding was then performed by applying One-Hot Encoding to variables such as \texttt{\{Marital status, Application mode, Course, \dots\}} via \texttt{pd.get\_dummies()}, and boolean values were converted to 0/1 integers using a \texttt{map} function. Following this, continuous numerical features like \texttt{\{Previous qualification (grade), Admission grade, \dots\}} were normalized to the \([0,1]\) range using \texttt{MinMaxScaler()}. The multi-class \texttt{Target} variable (“Enrolled”, “Dropout”, “Graduate”) was then converted to numeric labels (\(\{0\to \text{Dropout},\,1\to \text{Enrolled},\,2\to \text{Graduate}\}\)) with \texttt{LabelEncoder()}. Finally, the dataset was split into training, validation, and test sets using \texttt{train\_test\_split(..., stratify=y)} to preserve class proportions, allocating \(80\%\) of the data for training and validation (which was further split \(75\%/25\%\) respectively) and \(20\%\) for the final test set, ensuring stratified splitting throughout.

\subsection{Graph Construction via Clustering}
\label{subsec:graph_construction}
We explored two main strategies for constructing graphs from the tabular student data:

\noindent\textbf{Strategy 1: Silhouette Score Optimized Clustering:}
\begin{enumerate}
    \item \textbf{Dimensionality Reduction and Clustering Combinations}: We experimented with various combinations of dimensionality reduction techniques (\gls{pca}, \gls{umap}, \gls{tsne}, and \gls{pca} followed by \gls{umap} or \gls{tsne}) and clustering algorithms (K-Means and HDBSCAN).
    \item \textbf{Cluster Validation}: For each combination, the quality of the resulting clusters was evaluated using the Silhouette Score. This metric measures how similar an object is to its own cluster (cohesion) compared to other clusters (separation). Scores range from -1 to +1, with higher values indicating better-defined clusters.
    \item \textbf{Optimal Configuration Selection}: The combination yielding the highest Silhouette Score was selected for primary graph construction. In our experiments, this was \gls{umap} (to 10 components) followed by HDBSCAN clustering (with euclidean distance).
    \item \textbf{Graph Generation}: Nodes represent students. An undirected edge connects two student nodes if they are assigned to the same cluster by the HDBSCAN configuration. Outlier nodes would not form edges based on shared clusters or could be isolated. The original student features (after preprocessing) were used as node features.
\end{enumerate}

\noindent\textbf{Strategy 2: Fixed Configuration for GNN Benchmarking:}
\begin{enumerate}
    \item \textbf{Fixed Dimensionality Reduction}: Data was projected into a 2-dimensional space using \gls{pca} followed by \gls{umap}. This combination is often used for visualization and can reveal underlying manifold structures.
    \item \textbf{Graph Generation via Proximity}: A graph was constructed based on proximity in this 2D projected space. Edges were established between nodes (students) if they were among the \(k=5\) nearest neighbors of each other, identified using a KD-Tree. This creates a graph based on local similarity in the embedded space, rather than explicit cluster membership.
    \item Node features remained the original preprocessed student features.
\end{enumerate}
The properties of the resulting graphs, such as the number of connected components and graph density, were analyzed.

\subsection{Graph Neural Network Models}
\label{subsec:gnn_models}
We implemented two \gls{gnn} architectures using PyTorch Geometric \cite{fey2019fast}:

\noindent\textbf{1. Custom \Gls{gcn} Model:}
The architecture comprised three \gls{gcn} layers (\texttt{GCNConv}). Each convolutional layer was followed by \texttt{BatchNorm1d} for stabilization, a \texttt{LeakyReLU} activation function, and \texttt{Dropout} for regularization. The output from the final \gls{gcn} layer was passed through a fully connected layer to produce class predictions.

\noindent\textbf{2. GraphSAGE Model:}
This model also used three convolutional layers, but of type \texttt{SAGEConv}. Similar to the custom \gls{gcn}, each \texttt{SAGEConv} layer was followed by \texttt{BatchNorm1d} and \texttt{ReLU} activation, and \texttt{Dropout}. A final fully connected layer performed the classification.

\noindent\textbf{Training and Optimization:}
Both \gls{gnn} models were trained using the Adam optimizer \cite{dive} and Cross-Entropy Loss \cite{dive}. Hyperparameters (hidden dimension size, learning rate, dropout rate) were optimized using Optuna \cite{akiba2019optuna}, performing a specified number of trials. Early stopping was employed based on validation loss to prevent overfitting, with a patience of 15 epochs.

\subsection{Tabular Baseline Models}
\label{subsec:tabular_baselines}
To provide a strong comparative baseline, we trained and evaluated three established models directly on the preprocessed tabular data:
\begin{itemize}
    \item \textbf{\Gls{rf} and XGBoost}: Trained using 5-fold cross-validation. Hyperparameters  were optimized via random search over 5 combinations for each model, aiming to maximize performance on the validation folds. 

     \item \textbf{TabNet}: Model was configured with several key hyperparameters. These included \texttt{n\_d} and \texttt{n\_a} to set the widths of the decision prediction and attention embedding layers, respectively, within its attentive transformers. The number of sequential decision stages was defined by \texttt{n\_steps}, while \texttt{gamma} controlled feature selection sparsity. The architecture also incorporated \texttt{n\_independent} and \texttt{n\_shared} Gated Linear Unit (GLU) layers, applied independently to features and shared across features within GLU blocks, respectively. The \texttt{mask\_type} (e.g., \texttt{sparsemax} or \texttt{entmax}) dictated the feature selection masking strategy. Training utilized an Adam optimizer, a learning rate scheduler, and early stopping with a 20\% validation split, following a public Kaggle implementation\footnote{\url{https://www.kaggle.com/code/ar2197/tabnet-multi-class-and-binary-classification}}.   
    It was trained using an Adam optimizer, a learning rate scheduler, and early stopping based on validation performance (20\% validation split). \footnote{ Configured based on a public Kaggle implementation\url{https://www.kaggle.com/code/ar2197/tabnet-multi-class-and-binary-classification}}
\end{itemize}


\subsection{Evaluation Metrics}
\label{subsec:evaluation_metrics}
All models were evaluated on the held-out test set using standard classification metrics. For a multi-class classification problem with classes \(C_1, \dots, C_K\), let \(TP_k\), \(FP_k\), \(TN_k\), and \(FN_k\) denote the number of True Positives, False Positives, True Negatives, and False Negatives for class \(C_k\), respectively.

The \textbf{Accuracy} measures the overall proportion of correctly classified instances across all classes and is defined as:
\begin{equation}
\text{Accuracy} = \frac{\sum_{k=1}^{K} TP_k}{\text{Total number of instances}}
\end{equation}

For each class \(C_k\), \textbf{Precision}$_k$ quantifies the ability of the classifier not to label a sample as positive if it is negative for that class. It is calculated as:
\begin{equation}
\text{Precision}_k = \frac{TP_k}{TP_k + FP_k}
\end{equation}

Similarly, \textbf{Recall}$_k$ (or Sensitivity) for class \(C_k\) measures the ability of the classifier to find all positive samples belonging to that class:
\begin{equation}
\text{Recall}_k = \frac{TP_k}{TP_k + FN_k}
\end{equation}

The \textbf{F1-Score}$_k$ for class \(C_k\) is the harmonic mean of Precision$_k$ and Recall$_k$, providing a balance between them:
\begin{equation}
\text{F1-Score}_k = 2 \cdot \frac{\text{Precision}_k \cdot \text{Recall}_k}{\text{Precision}_k + \text{Recall}_k}
\end{equation}

To obtain an overall performance measure that accounts for all classes equally, especially given the potential class imbalance identified in Figure \ref{fig:class_distribution}, we also report macro-averaged versions of Precision, Recall, and F1-Score. Macro-averaging involves calculating the metric independently for each class and then taking the unweighted mean of these per-class scores. For example, Macro F1-Score is:
\begin{equation}
\text{Macro F1-Score} = \frac{1}{K} \sum_{k=1}^{K} \text{F1-Score}_k
\end{equation}
Macro Precision and Macro Recall are calculated analogously. These macro-averaged metrics are particularly important as they give equal weight to each class, regardless of its frequency.

\section{Experimental Setup}
\label{sec:experimental_setup}


\subsection{Dataset}

\begin{figure}[!ht]
    \centering
    \includegraphics[width=.8\textwidth]{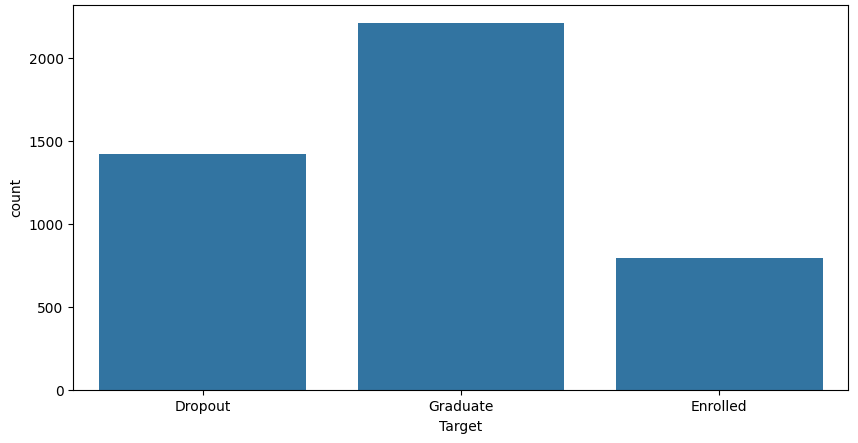} 
    \caption{Distribution of target classes in the student dataset} 
    \label{fig:class_distribution}
\end{figure}

After preprocessing, the dataset consisted of 4424 instances and 250 features. The target variable had three classes: ``Dropout'', ``Graduate'', and ``Enrolled''. As shown in Figure \ref{fig:class_distribution}, the dataset exhibits class imbalance, with the ``Graduate'' class being the most frequent, followed by ``Dropout'', and ``Enrolled'' being the least frequent. This imbalance was considered during model training and evaluation (focus on macro-averaged metrics).
%

\subsection{Implementation Details}
All experiments were conducted using \textbf{Python}, \textbf{Optuna} \cite{akiba2019optuna} for hyperparameter optimization and \textbf{NetworkX} \cite{hagberg2008exploring} for graph manipulation, along with \textbf{PyTorch Geometric}.
Hyperparameter tuning for \glspl{gnn} explored learning rates typically in the range \([10^{-4}, 10^{-1}]\), hidden dimensions from \(16\) to \(256\), and dropout rates from \(0.1\) to \(0.5\). For tabular models, ranges were set based on common practices for \gls{rf} and XGBoost.

\section{Results and Discussion}
\label{sec:results_discussion}

\subsection{Performance of Tabular Baselines}

The performance of the traditional \gls{ml} and \gls{dl} models on the tabular data is summarized in Table \ref{tab:tabular_results}.
\gls{rf} and XGBoost achieved competitive macro-averaged F1-scores (0.69 for XGBoost, 0.66 for RF) and overall accuracy (0.76). XGBoost demonstrated slightly better precision and recall balance. TabNet, the \gls{dl} baseline for tabular data, performed slightly lower, with a macro F1-score of 0.64 and accuracy of 0.73.
\begin{table}[!ht]
    \caption{Performance of Tabular Models}
    \label{tab:tabular_results}
    \centering
    \resizebox{.75\linewidth}{!}{
        \begin{tabular}{@{}llcccc@{}}
            \toprule
            Model          & Class/Avg     & Precision & Recall & F1-Score & Accuracy \\ \midrule
            \multirow{4}{*}{\parbox{0.1\linewidth}{Random Forest}} 
                & Dropout (0)   & 0.57      & 0.27   & 0.37     & \multirow{4}{*}{0.76} \\
                & Enrolled (1)  & 0.78      & 0.77   & 0.78     &          \\
                & Graduate (2)  & 0.78      & 0.93   & 0.85     &          \\
                & Macro Avg     & 0.71      & 0.66   & 0.66     &          \\ \midrule
            \multirow{4}{*}{XGBoost}       
                & Dropout (0)   & 0.52      & 0.42   & 0.46     & \multirow{4}{*}{0.76} \\
                & Enrolled (1)  & 0.78      & 0.75   & 0.76     &          \\
                & Graduate (2)  & 0.81      & 0.89   & 0.85     &          \\
                & Macro Avg     & 0.70      & 0.69   & 0.69     &          \\ \midrule
            \multirow{4}{*}{TabNet}        
                & Dropout (0)   & 0.48      & 0.29   & 0.36     & \multirow{4}{*}{0.73} \\
                & Enrolled (1)  & 0.77      & 0.70   & 0.73     &          \\
                & Graduate (2)  & 0.76      & 0.91   & 0.83     &          \\
                & Macro Avg     & 0.67      & 0.63   & 0.64     &          \\ \bottomrule
        \end{tabular}    
    }    
\end{table}

Notably, all tabular models struggled with the Dropout class (class 0), achieving F1-scores between 0.36 (TabNet) and 0.49 (XGBoost). This highlights the difficulty of predicting this specific class, likely due to its minority status or more complex patterns.

\subsection{Performance of GNN Models}
We evaluated the \gls{gnn} models on graphs constructed using the two proposed strategies.

\noindent\textbf{GNNs on Fixed PCA+UMAP Graph:}
Table \ref{tab:gnn_pca_umap_results} reports the results for this configuration. GraphSAGE achieved the best overall performance, with an accuracy of 0.7440 and a macro F1-score of 0.7583. In contrast, the custom GCN attained 0.7154 accuracy and a macro F1-score of 0.6884. 
\begin{table}[!ht]
\caption{GNN Performance on Fixed PCA+UMAP Graph - strategy 1}
\centering
\resizebox{.7\linewidth}{!}{
\begin{tabular}{lcccc}
\toprule
Model             & Accuracy & Precision & Recall & F1-Score \\ \midrule
GCN\_Gmod         & 0.7154   & 0.6941    & 0.7154 & 0.6884   \\
SAGE\_Gmod        & 0.7440   & 0.7880    & 0.7440 & 0.7583   \\ \bottomrule
\end{tabular}}
\label{tab:gnn_pca_umap_results}
\end{table}

\begin{figure}[!ht]
\centerline{\includegraphics[width=\columnwidth]{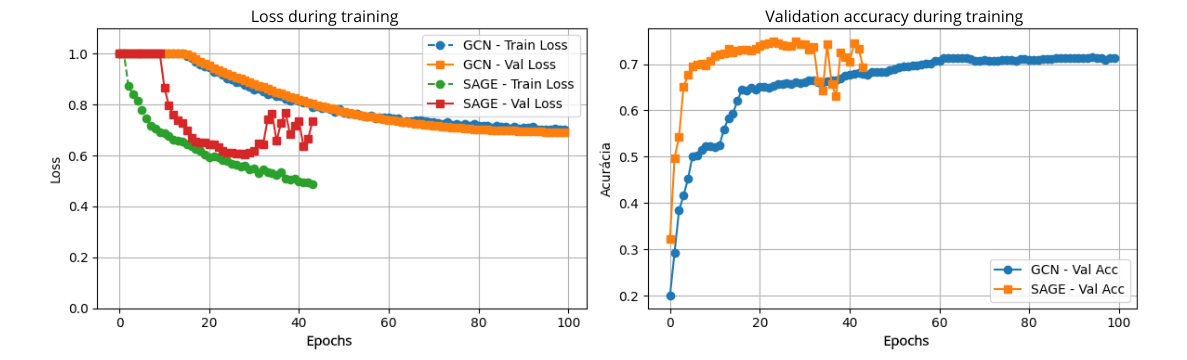}}
\caption{Training loss and validation accuracy for GNN models on strategy 1 (PCA+UMAP graph)}
\label{fig:gnn_pca_umap_curves}
\end{figure}

As shown in Figure \ref{fig:gnn_pca_umap_curves}, GraphSAGE drives its validation loss down very quickly, within the first 20 to 25 epochs, whereas the custom GCN only begins to settle around epoch 50. On the accuracy side, GraphSAGE peaks at roughly 0.74 – 0.75 by epoch 25 before exhibiting some oscillation, whereas the GCN gradually climbs and ultimately plateaus around 0.71 after roughly 60 epochs. Thus, GraphSAGE not only converges significantly faster but also achieves a modestly higher maximum validation accuracy compared to the custom GCN.

\begin{figure}[!ht]
\centerline{\includegraphics[width=\columnwidth]{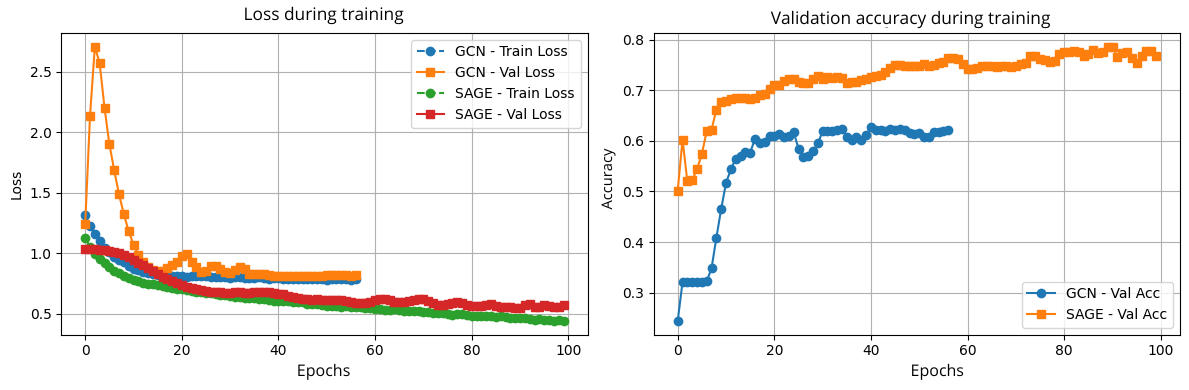}}
\caption{Training loss and validation accuracy for GNN models on strategy 2 (UMAP(10) + HDBSCAN graph)}
\label{fig:umap_hdbscan_curves}
\end{figure}

\noindent\textbf{GNNs on Silhouette-Optimized UMAP (10) + HDBSCAN Graph (\(G\)):} 
Table \ref{tab:gnn_pca_kmeans_results} and Figure \ref{fig:umap_hdbscan_curves} present results for the graph constructed using UMAP (10) and HDBSCAN (Silhouette Score: 0.8256). Again, GraphSAGE outperformed GCN, achieving an accuracy of 0.7756 and a macro F1-score of 0.7658. The GCN reached 0.6265 accuracy and a 0.5565 F1-score. This GraphSAGE result represents the best performance among all models tested.
\begin{table}[!ht]
\caption{GNN Performance on Silhouette-Optimized UMAP (10) + HDBSCAN Graph - strategy 2}
\centering
\resizebox{.65\linewidth}{!}{
\begin{tabular}{@{}lcccc@{}}
\toprule
Model             & Accuracy & Precision & Recall & F1-Score \\ \midrule
GCN\_G            & 0.6265   & 0.6041    & 0.6265 & 0.5565   \\
SAGE\_G           & 0.7756   & 0.7747    & 0.7756 & 0.7658   \\ \bottomrule
\end{tabular}}
\label{tab:gnn_pca_kmeans_results}
\end{table}

\subsection{Comparative Analysis and Discussion}

Table \ref{tab:overall_comparison} provides an overall comparison. Among all configurations, the GraphSAGE model trained on the UMAP(10)+HDBSCAN graph achieved the highest macro F1-score (76.58\%) and accuracy (77.56\%).

\begin{table}[!ht]
\caption{Overall Model Performance Comparison}
\centering
\resizebox{\columnwidth}{!}{%
\begin{tabular}{@{}lcccc@{}}
\toprule
Model                          & Accuracy   & Precision  & Recall     & F1-Score   \\ \midrule
Random Forest (ML Tabular)     & 76.16\%    & 74.15\%    & 76.16\%    & 73.88\%    \\
XGBoost (ML Tabular)           & 76.05\%    & 74.95\%    & 76.05\%    & 75.30\%    \\
TabNet (DL Tabular)            & 73.11\%    & 66.97\%    & 63.28\%    & 64.01\%    \\ \midrule
GCN (UMAP (10) + HDBSCAN)      & 62.66\%    & 60.41\%    & 62.66\%    & 55.65\%    \\
GraphSAGE (UMAP (10) + HDBSCAN)& \textbf{77.56\%} & 77.47\%    & \textbf{77.56\%} & \textbf{76.58\%} \\ \midrule
GCN (PCA + UMAP)               & 71.54\%    & 69.41\%    & 71.54\%    & 68.84\%    \\
GraphSAGE (PCA + UMAP)         & 74.40\%    & \textbf{78.80\%} & 74.40\%    & 75.83\%    \\ \bottomrule
\end{tabular}
}
\label{tab:overall_comparison}
\end{table}

\noindent\textbf{Addressing Research Questions:}

\begin{figure}[!ht]
  \centering
  \includegraphics[width=0.6\textwidth]{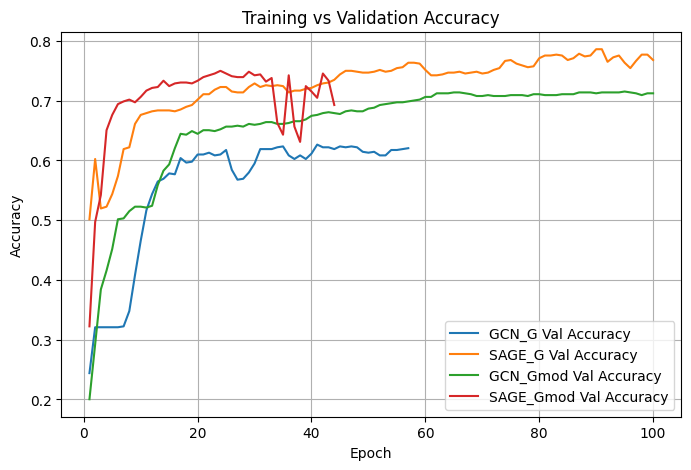}
  \caption{Training versus Validation Accuracy over 100 Epochs. GCN\_G and SAGE\_G use UMAP(10) + HDBSCAN; GCN\_Gmod and SAGE\_Gmod use PCA + UMAP.}
  \label{fig:train_val_accuracy}
\end{figure}

\begin{figure}[!ht]
  \centering
  \includegraphics[width=0.6\textwidth]{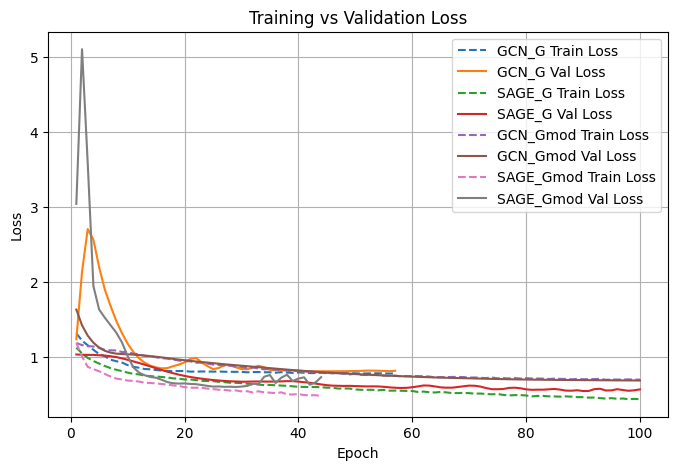}
  \caption{Training versus Validation Loss over 100 Epochs. GCN\_G and SAGE\_G use UMAP(10) + HDBSCAN; GCN\_Gmod and SAGE\_Gmod use PCA + UMAP.}
  \label{fig:train_val_loss}
\end{figure}

\noindent\textbf{RQ1 (Can graph conversion improve prediction?):} Yes, but only modestly. GraphSAGE trained on the UMAP(10)+HDBSCAN graph achieves the highest validation accuracy (around 0.78) and lowest validation loss ($\approx$0.8) after convergence, as seen in Figures \ref{fig:train_val_accuracy} and \ref{fig:train_val_loss}. This configuration yields a macro F1-score of 76.58\% (accuracy 77.56\%), roughly 1.3 percentage points higher than the best tabular baseline, XGBoost (75.30\% F1, 76.05\% accuracy). Other GNN variants—such as GCN on UMAP+HDBSCAN—suffer from slower accuracy growth and higher validation loss (e.g., GCN\_G peaks near 0.62 accuracy), while GraphSAGE on PCA+UMAP (SAGE\_Gmod) converges to an F1-score of 75.83\% ($\approx$ 0.71 accuracy). In short, although graph-based models can outperform tabular ones, the gain is relatively small and strongly depends on both architecture and graph construction strategy.

\noindent\textbf{RQ2 (Impact of graph-construction strategy?):} The graph-construction method has a decisive impact on GNN performance. UMAP(10)+HDBSCAN produces a graph in which GraphSAGE (SAGE\_G) quickly reaches a stable validation accuracy above 0.75 and validation loss below 1.0 by epoch 20, whereas PCA+UMAP (SAGE\_Gmod) converges more gradually to around 0.71 accuracy and 0.75 loss by epoch 100 (Figures \ref{fig:train_val_accuracy}–\ref{fig:train_val_loss}). In both embedding strategies, GraphSAGE outperforms our custom GCN: GCN\_G (UMAP+HDBSCAN) flattens around 0.62 accuracy and suffers higher loss, and GCN\_Gmod (PCA+UMAP) plateaus near 0.70 accuracy with loss $\approx$ 0.9. These dynamics highlight that higher-quality clustering and dimensionality reduction (UMAP + HDBSCAN) yield more discriminative graph structures, and that GraphSAGE’s sampling/aggregation mechanism is more robust on these generated graphs compared to our GCN.

\section{Conclusion}
\label{sec:conclusion}

This study evaluated the conversion of tabular student data into graph representations for dropout prediction using various GNN architectures and graph‐construction methods. A GraphSAGE model applied to a graph built via UMAP (10 dimensions) followed by HDBSCAN clustering achieved the highest performance, with an F1‐score of 76.58\%. This slightly surpasses the best tabular baseline (XGBoost, F1 = 75.30\%) and outperforms Random Forest (F1 = 73.88\%) and TabNet (F1 = 64.01\%). Even when using PCA + UMAP for graph construction, GraphSAGE reached an F1 of 75.83\%, demonstrating that appropriately engineered graphs allow GNNs to match or exceed classical ML models on tabular features.

In contrast, GCNs performed substantially worse—achieving only 55.65\% F1 with UMAP (10) + HDBSCAN and 68.84\% F1 with PCA + UMAP—highlighting the importance of both the GNN architecture and the graph‐generation strategy. Overall, GraphSAGE with UMAP + HDBSCAN stands out as the most promising approach by delivering balanced precision and recall (77.47\% and 77.56\%, respectively) and improving macro F1 over XGBoost. Future work should explore more sophisticated graph‐construction techniques—such as learnable edges or domain‐informed connectivity—together with alternative GNN architectures and pre‐training schemes to further enhance dropout‐prediction performance.

\section*{Acknowledgment}
This work was supported by the Conselho Nacional de Desenvolvimento Científico e Tecnológico (CNPq, grants 308400/2022-4, 307151/2022-0), Coordenação de Aperfeiçoamento de Pessoal de Nível Superior (CAPES - grant 001), Fundação de Amparo à Pesquisa do Estado de Minas Gerais (FAPEMIG, grant APQ-01647-22). We also thank the Universidade Federal de Ouro Preto (UFOP) for their support.





\bibliographystyle{main}  
\bibliography{main}

\end{document}